# Producing NLP-based On-line Contentware

**Francis Wolinski** [1]    **Frantz Vichot** [1]    **Olivier Grémont** [2]

[1] Informatique CDC/DTA, 94114 Arcueil, France
[2] Informatique CDC/GIRET, 33059 Bordeaux, France
{fwolinski, fvichot, ogremont}@icdc.caissedesdepots.fr

## Abstract

For its internal needs as well as for commercial purposes, CDC Group has produced several NLP-based on-line contentware applications for years. The development process of such applications is subject to numerous constraints such as quality of service, integration of new advances in NLP, direct reactions from users, continuous versioning, short delivery deadlines and cost control. Following this industrial and commercial experience, malleability of the applications, their openness towards foreign components, efficiency of applications and their ease of exploitation have appeared to be key points. In this paper, we describe TalLab, a powerful architecture for on-line contentware which fulfils these requirements.

**Keywords:** On-line Contentware, NLP-based Industrial Application, Language Engineering, Software Architecture, Multi-Agents System

## 1. Introduction

The recent period has seen the great expansion of the worldwide internet and of companies' intranets. At the same time, the mass of available electronic documents has grown astronomically. The result is a large demand from users for on-line contentware. These applications are available on networks and presents documents with a certain added value (e.g., text categorization, information retrieval, text mining, hypertext navigation). Nowadays, as proved by many works, see (AAAI 97) for instance, NLP techniques are mature enough to become a major means to bring such added value automatically to electronic documents. Considering this kind of application as strategic, Caisse des dépôts et consignations (CDC) has an experience of several years in the producing of NLP-based on-line contentware.

However, producing such complex applications, broadcast on a wide network, consulted by many users and based on NLP advanced techniques, requires taking into account numerous Industrial and Commercial (I&C) constraints: quality of service, integration of new advances in NLP, direct reactions from users, continuous versioning, short delivery deadlines and cost control. Moreover, these applications have to be deployed on machines and networks of organizations and have to be monitored daily by operators who are not language engineers.

Therefore, the crucial question is what kind of software architecture should be used to manage such complexity? Some authors have noticed that the shift from Computational Linguistics to Language Engineering gave a new impetus to architectures that are dedicated to NLP (Cunningham et al. 96b; Zajac et al. 97). As a result, some powerful architectures have been proposed. Their aim is to facilitate the work of R&D language engineers who test new linguistic theories and who improve NLP components. A next step is to make these high-tech applications be a standard part of organizations' information systems.

In this paper, we report our experience in producing large NLP-based on-line applications. For that purpose, we have implemented a powerful architecture named TalLab. Its aim is to ease the work of software engineers producing, deploying and monitoring the applications. In the design process of TalLab, the set of requirements listed by existing architectures has had to be completed with additional I&C requirements.

Section 2 gives an overview of NLP-based on-line contentware produced by CDC. Section 3 lists the set of requirements for producing such applications. Section 4 discusses some well-established choices of NLP architectures that have had to be reconsidered in the design of TalLab. Section 5 gives the design guidelines of TalLab and finally Section 6 describes its salient technical features.

## 2. A strong experience of NLP-based on-line contentware

For its internal needs as well as for commercial purposes, CDC has produced several NLP-based on-line contentware applications.

Since 1993, a real-time information application has been available on CDC's intranet (Fig. 1). It performs a classification of the economic data flow from Agence France-Presse (A.F.P., the French news agency) into 100 or so topics for intranet users such as competitive intelligence watchers, financial analysts, fund managers, documentalists and executives.

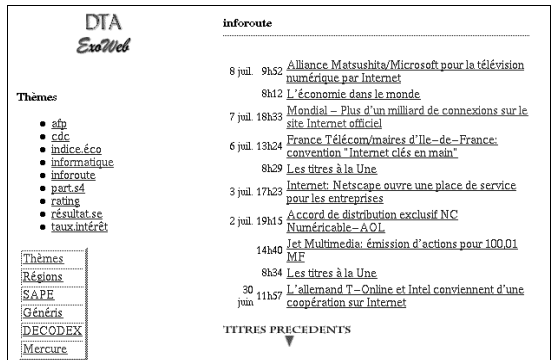

Figure 1: Filtering A.F.P. economic flow

In 1995, this real-time application was connected to the SAPE decision support system. This system provides ways to manage the complex information of companies' shareholders and gives the capability to simulate moves in the composition of a firm's shareholders. In order to enable fund managers to feed SAPE's knowledge base, information are automatically extracted from news that are dealing with shareholdings (Vichot et al. 99).

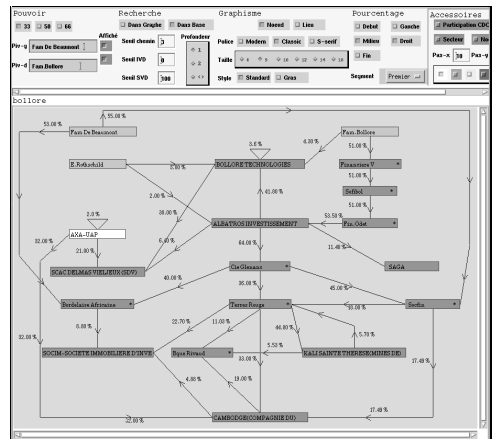

Figure 2: Feeding an expert system

In 1997, CDC-Mercure, a new subsidiary of CDC, launched a commercial internet application (see `http://www.cdc-mercure.fr`) named « Mercure: the daily internet newspaper for local authorities » (Fig. 3). It provides a classification of A.F.P. news into 80 different topics for managers in local authorities (1,300 paying clients, 2,000 planned by the end of 1998).

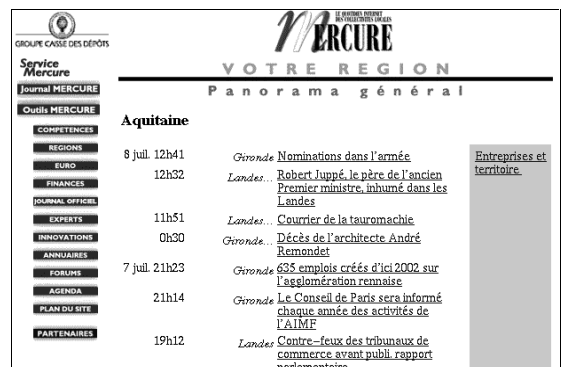

Figure 3: Commercial internet application

Finally, an on-going application, based on a prototype described in (Vichot et al. 97), provides a hypertext navigation on a large database of A.F.P. news (Fig. 4). The idea is to enable users to explore a huge documents base including several months or years of A.F.P. news. This hypertext is built by automatically generating the hyperlinks thanks to information extraction techniques. For example, named entities and related events as well are as many entry points for an intelligent browsing.

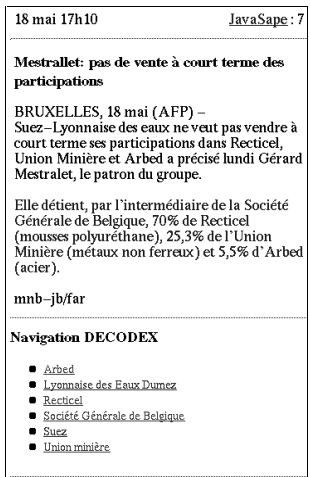

Figure 4: Hypertext navigation

# 3. Requirements for NLP- based on-line contentware

Motivating requirements for NLP architecture is not a new concern. For instance, (Cunningham et al. 96b) have already listed « a set of requirements for the provision of software infrastructure for NLP R&D ». In this section, we complete them with an I&C point of view.

## 3.1 Providing malleability to facilitate the evolution of applications

With on-line contentware, users appreciate direct reaction and continuous versioning. More than in usual software development, on-line contentware needs an iterative development process with many specification-implementation-evaluation loops. Moreover, the pressure from customers and the availability of new NLP techniques or software components lead to building up prototypes in order to propose improvements or extensions to existing applications. The ease of the development of full-scale demonstrations is very important.

## 3.2 Maximizing openness to take advantage of foreign components

Openness has appeared to be essential because it governs the reuse of existing know-how, thanks to available components, and therefore the feasibility of high value-added applications within short delivery deadlines and *reasonable costs*. The central question is the interoperability of software components which have been developed by different teams in different languages and using different implementation techniques. The crucial point is that foreign components must be integrated with as low as possible adaptation efforts.

## 3.3 Increasing efficiency to extend the scope of applications

Natural language processing is known to make intensive use of computer resources especially in CPU. Efficiency is an essential constraint because of the large flows[1] and of the high volumes[2] of documents that our applications do process. To optimize the available resources, applications have to be easily distributed. This must be done by supporting concurrent architectures to carry out parallel processing (i.e. one document is processed at the same time by several modules) as well as pipe-line architectures to carry out sequential processing (i.e. several documents are processed at the same time by different modules).

## 3.4 Ensuring exploitability to guarantee the integration of applications

With the development of the internet/intranet, on-line contentware is to be considered as a standard application. Three main problems have to be considered:
- *Deployability*: applications should be easily installed and configured on any standard Management Information System (MIS) whatever its own architecture.
- *Reliability*: in case of a component crash, software protection, knowledge integrity and process conservation of applications should be ensured.
- *Controllability:* applications should be easily monitored by operators who know very little about NLP and easy connection to standard supervision tools packages, such as EcoTOOLS or SunNet Manager, should also be enabled.

---

[1] An average of 55 news/hour (110 Kb/h) up to 200 news/hour (650 Kb/h).
[2] A complete year of A.F.P. news amounts to 400,000 news or 800 Mb.

## 4. R&D language engineering architectures

Regarding the knowledge exchanged between components, NLP systems generally rely on a single format which is considered as a standard: MULTEXT in ALEP (Simkins 94), TIPSTER in GATE (Cunningham et al. 96a) and in Corelli (Zajac et al. 97), SGML in LT-NSL (McKelvie et al. 97), CG in Caramel (Vapillon et al. 97), etc.

Obviously in NLP, the multiplication of such « standards » associated to different linguistics approaches does not facilitate neither reuse of components nor openness of systems. For example, to integrate a new component in ALEP, the component's own formalism must be first translated into a neutral one which is supplied by this platform. This overhead is too heavy a investment for testing foreign components.

Moreover, the use of a single format generally leads to the implementation of dedicated APIs. They do not facilitate neither reuse nor openness, as well. A good example is GATE, a complete R&D architecture which is dedicated to language engineering. GATE relies on a collection of specific wrappers. Hence, the integration of a component requires to master well its implementation. Conversely, the standardization of components interface increases malleability and controllability of applications.

In the same idea, Corelli provides a software architecture for reuse and integration of heterogeneous NLP components. Their solution relies on a specific component integration API based on CORBA. It involves precisely mastering the components to integrate, as well. In fact, Corelli's developers noticed that in some cases this solution is not suitable and they had to resort to basic system calls.

This very solution is generalized in Kasuga, a speech translation blackboard prototype, where (Boitet & Seligman 94) introduced the idea of managers that encapsulate NLP components and set up asynchronous communications with a coordinator. Such a technique facilitates the integration of new components but it requires whenever to modify the coordinator. In other respects, such a central scheduler, that is requested by every component, is a source of runtime inefficiency and of software weakness.

## 5. TalLab design guidelines

According to the requirements expounded in Section 3 and the experiences of R&D systems which are summed up in previous section, our implementation of TalLab has followed three design guidelines:
- relying on a multi-agents system,
- reusing the operating system wherever it is possible,
- refusing to impose a single standard for components implementation.

### 5.1 A multi-agents system …

In the field of distributed artificial intelligence (Genesereth & Ketchpel 94) have shown in detail how interoperability of heterogeneous software components can be tackled with multi-agents techniques. Moreover, agents communicate in an asynchronous mode which is a much more robust way than others (pipes, client/server, linkage). According to the agent model described in (Guessoum & Briot 98), an agent is made of three elements: a behavior, an activity and an address. In TalLab, each instance of a software component is integrated as an asynchronous agent. The NLP component itself represents the behavior. The wrapper that encapsulates the component and ensures its integration in the whole system stands for the activity. The address consists of a message box in which processed documents ids are queued. Finally, TalLab agents feature a persistence function that stores useful information.

### 5.2 … embedded in the operating system …

TalLab is embedded in the operating system. Reusing the OS is directly inspired from blackboard architectures. On the one hand, in AI the use of knowledge sources contributing to solve a problem shares the same purposes as the modularity introduced in NLP. On the other hand, (Engelmore & Morgan 88) noticed that « the control aspects of a blackboard have much in common with operating systems design ». So, resorting to the OS is an important choice. TalLab delegates the OS to perform the complex task of integrating heterogeneous software components. This allows to reuse directly a huge, efficient and reliable amount of code.

### 5.3 … not imposing a single standard

In TalLab, no standard of implementation is imposed on software components. This point is valid for operating systems as well as for programming languages, for communication protocols as well as for knowledge formats, etc. For instance, despite the great efforts made in standardization, we have seen in Section 4 that several standards of knowledge formats are still in competition. Choosing one specific format would involve adapting existing components to that format and possibly future components, even for a simple evaluation. In our applications, we noticed that, first, proprietary formats are compact and well-adapted to their corresponding component, and second, that the knowledge produced by a component is generally used by a small number of others. Therefore, we decided to integrate each component with its own particularities with a low adaptation cost.

## 6. TalLab salient technical features

This section describes the salient technical features of TalLab. It shows how the guidelines used to design TalLab and presented in previous section make it fulfilling the I&C requirements expressed in Section 4.

### 6.1 Building circuits of value-added agents

Agents that build up an application are explicitly organized in circuits which form oriented acyclic graphs of producers-suppliers of knowledge (Fig. 5). Documents are processed by agents in flows that follow this graph exactly.

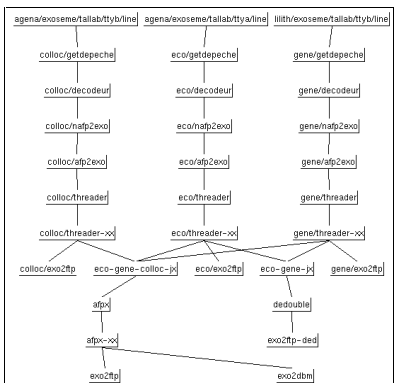

Figure 5: A circuit of agents

Each agent can use the knowledge produced by previous agents and generates knowledge that can be used by next agents. For a root agent, the initial knowledge of a document is its text, and possibly timestamp or other extra-textual information. The whole application can be seen as a circuit of value-added agents which provides by emergence interesting knowledge about documents.

With such circuits of agents, TalLab reaches malleability. The programmer may create and plug a new branch of a main application to make a full-scale evaluation of a new technique. For instance, two new applications – implemented as circuits – have been added to Mercure: statistical clustering of news using the inverse document frequency factor and indexation of news using Search'97, the search engine from Verity, Inc.

### 6.2 Using adapters to transform NLP components into agents

Each instance of a software component is encapsulated in an adapter using the *transducers* technique described in (Genesereth & Ketchpel 94). The latter simply transforms a component into an asynchronous agent. Such agents are all designed with a uniform software activity. The flow of documents ids to be processed by an agent is controlled by a pooling system performed by the adapter and is implemented in the file system as a FIFO shared queue. Each agent is put in its own directory and any information produced by the agent or by its underlying component is locally stored (e.g., the flow of processed documents ids, information about its own process, the produced knowledge). So this information can be accessed either by the agent itself, or by other agents, or again by any software tool.

Implemented in Perl, called « a chewing gum and baling wire language » by its own designer, adapters are easy-to-write wrappers. They enable TalLab to be open to foreign components at the process level. All NLP modules have been effectively integrated in TalLab in this way: the Sylex syntactic analyzer (Constant 91), a semantic analyzer (Landau et al. 93), a proper names gazetteer (Wolinski et al. 95), a statistical matcher (Treihlou 96), etc. Recently, the integration of Search'97 was done in only few days by implementing such an adapter.

### 6.3 Providing a library of transistor-like agents

Besides usual NLP components, TalLab includes a library of *transistor-like agents*. For instance, *multiplexers* duplicate a document flow into several identical ones; *synchronizers* merge several document flows into a unique one; *dispatchers* split a document flow into several ones; *filters* map a document flow with a boolean function onto another one; *networkers* transmit document flows or data across network proxies or internet firewalls using the FTP protocol; etc.

If usual NLP components introduce some intelligence in the agents, such tiny agents introduce some intelligence in the whole circuits. First, such agents facilitate the malleability of applications by providing simple means to create new branches of a main application. Second, they enable to optimize the use of resources by distributing applications. For instance, combining a dispatcher with networkers may carry out concurrent as well as pipeline processing. Third, they facilitate the deployability of applications by exactly adapting the applications to the target MIS. For instance, specific networkers have been implemented to go across proxies[3] and internet firewalls[4]. For example, commercial application Mercure has had to be deployed on three distinct networks: the technical network comprising A.F.P. satellite dishes, the internal network where standard programs run and the internet network on which the application is available to end-users.

### 6.4 Using translators to deal with format incompatibilities

To decrease the adaptation costs of foreign components, each component is integrated with its own particularities. Consequently, components may produce knowledge in different and incompatible formats. Additional agents named *translators* solve this problem by converting the format of a component into another.

In a multi-agents system, the resort to translators appears therefore to be a natural and efficient means to ensure the openness of applications at the data level. For instance, the integration of Search'97 was done by adjoining a HTML translator to the corresponding circuit so that news items can be indexed by the search engine.

### 6.5 Managing the modularity of applications

Starting or stopping a whole application can be a complex task for an operator. This is due to the modularity, which multiplies the number of processes, and to the distribution, which multiplies the number of machines. The homogeneity of adapters enables the writing of simple procedures to perform these tasks. The mastering of this complexity is also achieved by each adapter itself. For instance, when an agent is activated, it begins by reading the information stored at the time of its very last activation and by performing an OS instruction, that kills the corresponding process on the machine on which it had been launched. In consequence, the very same agent cannot be launched twice. These facilities take part in the exploitability of applications. So operators do not have to manage this information individually for each agent. Moreover, in order to have applications being easily monitored by operators, a connection to EcoTOOLS from Compuware Corp. has been enabled.

### 6.6 Recovering from a component crash

Adapters implement a mechanism for dealing with component crashes. When processing a document, an agent simply stores the document's id and the number of attempts for dealing with it. Over a fixed threshold, the document is purely and simply skipped. In this way, the component may continue to process the next documents and information about the unprocessable document are kept for later debugging. Dealing with the unavoidable component crashes, due to large corpora processed by our applications, improves their reliability. In consequence, applications are implemented in a way that avoids complex debugging process by operators who in any case are not language engineers and know very little about NLP.

---

[3] Inter-network connection machines.

[4] Proxies enforcing access control policies.

## 7. Conclusion and perspectives

This paper has reported our experience in building an architecture dedicated to produce on-line applications which rely on many NLP components. The industrialization and the commercialization contexts have lead us to deal with specific difficulties, for which literature does not give appropriate solutions. TalLab was designed to solve applications development problems, such as malleability and openness, as well as applications operating problems, such as efficiency and exploitability. Its main contribution lies in *an OS-based multi-agents system relying on an electronic component assembly model*. It is responsible for its simplicity and its power.

However, instead of Kasuga or Corelli, the scope of TalLab is still limited to applications that are mappable to agents organized in an oriented acyclic graph. More complex applications may need a closer cooperation between components. We are working on a special agent able to deal with this kind of strong interaction.

## References

(AAAI 97) *Working Notes from the AAAI Spring Symposium on Natural Language Processing for the World Wide Web*, Stanford, USA, 1997.

(Boitet & Seligman 94) Ch. Boitet, M. Seligman, *The Whiteboard Architecture: a Way to Integrate Heterogenous Components of NLP Systems*, International Conference on Computational Linguistics, COLING'94, Kyoto, Japan, 1994.

(Constant 91) P. Constant, *Analyse syntaxique par couche*, Thèse Télécom Paris, France, 1991.

(Cunningham et al. 96a) H. Cunningham, Y. Wilks and R. Gaizauskas, *GATE - A General Architecture for Text Engineering*, International Conference on Computational Linguistics, COLING'96, Kyoto, Japan, 1996.

(Cunningham et al. 96b) H. Cunningham, Y. Wilks and R. Gaizauskas, *New methods, Current Trends and Software Infrastructure for NLP*, Conference on New Methods in Natural Language Processing, NeMLaP'96, Ankara, Turkey, 1996.

(Engelmore & Morgan 88) R. Engelmore, T. Morgan, *Blackboard Systems*, Addison-Wesley, 1988.

(Genesereth & Ketchpel 94) M. Genesereth, S. Ketchpel, *Software agents*, Intelligent Agents, Communications of the ACM, New York, USA, 1994.

(Guessoum & Briot 98) Z. Guessoum, J.-P. Briot, *From Active Objects to Autonomous Agents*, Rapport LiP6, 98-015, Université de Paris 6, Paris, France, 1998.

(Landau et al. 93) M.-C. Landau, F. Sillion and F. Vichot, *Exoseme: a Thematic Document Filtering System*, Intelligence Artificielle'93, Avignon, France, 1993.

(McKelvie et al. 97) D. McKelvie, Ch. Brew and H. Thompson, *Using SGML as a Basis for Data-Intensive NLP*, Applied Natural Language Processing, ANLP'97, Washington, USA, 1997.

(Simkins 94) N. Simkins, *ALEP: An open Architecture for Language Engineering*, Language Engineering Convention, Paris, France, 1994.

(Treilhou 96) Ph. Treilhou, *Application des techniques d'inférence statistique à la gestion d'une base de connaissances utilisée en traitement automatique du langage*, Rapport de DEA, INRA, France, 1996.

(Vapillon et al. 97) J. Vapillon, X. Briffault, G. Sabah and K. Chibout, *An Object-Oriented Linguistic Engineering Environment using LFG and CG*, Computational Environments for Grammar Development and Linguistic Engineering, Association for Computational Linguistics, Workshop ACL'97, Madrid, Spain, 1997.

(Vichot et al. 97) F. Vichot, F. Wolinski, J. Tomeh, S. Guennou, B. Dillet and S. Aidjan, *High Precision Hypertext Navigation Based on NLP Automatic Extractions*, Hypertext Information Retrieval Multimedia, HIM'97, Dortmund, Germany, 1997.

(Vichot et al. 99) F. Vichot, F. Wolinski, H.-C. Ferri and D. Urbani, *Using Information Extraction for Knowledge Entering*, in S. Tzafestas (Ed.), *Advances in Intelligent Systems: Concepts, Tools and Applications*, Kluwer Academic Publishers, 1999, to appear.

(Wolinski et al. 95) F. Wolinski, F. Vichot and B. Dillet, *Automatic Processing of Proper Names in Texts*, European Chapter of the Association for Computational Linguistics, EACL'95, Dublin, Ireland, 1995.

(Zajac et al. 97) R. Zajac, M. Casper and N. Sharples, *An Open Distributed Architecture for Reuse and Integration of Heterogeneous NLP Components*, Applied Natural Language Processing, ANLP'97, Washington, USA, 1997.